\documentclass[letterpaper, 10 pt, conference]{ieeeconf}  
\usepackage{xcolor}

\usepackage{cite}
\usepackage{graphicx}
\usepackage{eurosym}
\usepackage{caption}
\usepackage{subcaption}
\usepackage{array}
\usepackage{seqsplit}
\usepackage{hyperref}
\newcolumntype{L}[1]{>{\raggedright\arraybackslash}p{#1}}
\usepackage{tikz}
\newcommand*\circled[2][2pt]{%
  \tikz[baseline=(char.base)]{%
    \node[draw,circle,inner sep=#1] (char) {\strut #2};}}

\IEEEoverridecommandlockouts                              

\title{\LARGE \bf RODEO: RObotic DEcentralized Organization}

\author{Milan Groshev, Eduardo Castell{\'{o}} Ferrer\\
\normalsize{CyPhy Life, Robotics \& AI Lab, School of Science \& Technology, IE University, Spain}}

\begin{document}
\maketitle
\urlstyle{same}
\thispagestyle{empty}
\pagestyle{empty}

\begin{abstract}
Robots are improving their autonomy with minimal human supervision. However, auditable actions, transparent decision processes, and new human-robot interaction models are still missing requirements to achieve extended robot autonomy. To tackle these challenges, we propose RODEO (RObotic DEcentralized Organization), a blockchain-based framework that integrates trust and accountability mechanisms for robots. This paper formalizes Decentralized Autonomous Organizations (DAOs) for service robots. First, it provides a ROS–ETH bridge between the DAO and the robots. Second, it offers templates that enable organizations (e.g., companies, universities) to integrate service robots into their operations. Third, it provides proof-verification mechanisms that allow robot actions to be auditable. In our experimental setup, a mobile robot was deployed as a trash collector in a lab scenario. The robot collects trash and uses a smart bin to sort and dispose of it correctly. Then, the robot submits a proof of the successful operation and is compensated in DAO tokens. Finally, the robot re-invests the acquired funds to purchase battery charging services. Data collected in a three day experiment show that the robot doubled its income and reinvested funds to extend its operating time. The proof-validation times of approximately one minute ensured verifiable task execution, while the accumulated robot income successfully funded up to 88 hours of future autonomous operation. The results of this research give insights about how robots and organizations can coordinate tasks and payments with auditable execution proofs and on-chain settlement. 
\end{abstract}

\section{INTRODUCTION}

As robotics technology advances and costs fall, robots are becoming increasingly common in our society~\cite{Kusnirakova2025}: from autonomous vehicles~\cite{Badue2021} and delivery drones~\cite{Sorbelli2024} to service robots in hospitals, homes, and industry~\cite{IFR2022}. As these systems proliferate, there is increasing pressure to develop methods that extend their autonomy, enabling robots to operate reliably with minimal human supervision in complex, dynamic, and uncertain environments~\cite{DARPA2017}. To support this, research is pushing forward in areas such as perception, adaptive control, online learning, fault detection, and decentralized coordination~\cite{Bossens2022}. These capabilities are essential if robots are to be trusted in daily tasks and to scale to real-world settings.

However, with increased autonomy comes the need for transparency and robot behavior explainability~\cite{Winfield2017}. Autonomous systems can manifest complex emergent behaviors, particularly when multiple agents interact, or when learning-based modules are involved~\cite{castello2016adaptive}. Without mechanisms for verification, auditing, and reliability, robots are unlikely to win public trust~\cite{Winfield2017}. Prior work has demonstrated architectures for accountability and explainability using logging, formal verification, or runtime monitoring to ensure that robot actions can be traced and explained~\cite{Mayoral2018}. Despite this prior work, no single framework addresses all current challenges in the field.

In the meantime, blockchain technology has emerged as a promising framework for digital trust, offering a distributed ledger with immutable audit trails~\cite{Castello-trust}. The programmable nature of blockchain enables Decentralized Autonomous Organizations (DAOs), where operational rules and decision making are encoded on chain. These attributes make blockchain technology a potential candidate to create systems that require autonomy, trust, transparency, and tamper‐resistance. 

In the robotics field, blockchain has been explored for robot event logging~\cite{ferrer2018robochain} , robot-to-robot interactions~\cite{danilov2018robonomics}  and  decentralized market place for robotic capabilities~\cite{peaq_machine_economy}. Recent state of the art efforts -- discussed in detail in Sec.~\ref{sec:soa} --  started combining DAOs and robotics to build coordination layers and frameworks that treat machines as independent agents~\cite{openmind2023fabric}, ~\cite{symbiotic2024sharedsecurity}. However, despite these advancements current blockchain solutions typically focus on data logging, general coordination and orchestration, rather than providing a organizational stack that incorporates service robots into daily operations. For robots to participate in blockchain-based organizations (e.g., by performing physical tasks) we require mechanisms that bridge digital trust with robotic embodiment, actuation, and sensing~\cite{Castello-trust}.

To address this challenge, we propose RODEO (RObotic DEcentralized Organization): a framework for robot-driven organizations (i.e., an organization where a robot can perform tasks as part of its operation) that users can customize in terms of tasks and service definitions. RODEO ties together blockchain-based trust and accountability with embodied robots, providing interfaces and control mechanisms so that organizations can specify what is done, how it is done, and how it is verified. In this paper, we first formalize DAOs for service robots. Second, we develop a ROS–ETH bridge that exposes task, service and proofs between the robot and blockchain environments. Third, we propose a DAO template designed to allow organizations to integrate robots in their daily activities. Finally, we evaluate and discuss the results of a proof-of-concept setup where a robot performs an action, generates income, reinvesting it in its own autonomy. Our aim is that the insights provided in this paper set the foundation for future real-world robot-driven DAOs.

\section{Related Work}
\label{sec:soa}
Blockchain has recently attracted significant attention in robotics as a means to support decentralized coordination, trust, and accountability. Surveys highlight applications in multi-robot collaboration, swarm security, task verification, and service marketplaces~\cite{srinivas2021survey}. Specific projects such as RoboChain~\cite{ferrer2018robochain} shows that blockchain can log robot actions immutably or provide secure interfaces to SC (i.e., self-executing digital agreements whose terms are written directly into the blockchain). Along these lines, the first ROS-Ethereum interface~\cite{zhang2022rosethereum} supports event logging and basic calls, yet a standardized interface that lets robots join and act within blockchain-based ecosystems is still missing. In short, current systems show that combining robotics with blockchain-based technology is feasible, but they stop at logging or low-level integrations (e.g., for conducting data-oriented analysis), rather than delivering an organizational stack that integrates service robots into day to day operations.

\begin{figure*}[t]
  \centering
  \includegraphics[width=\textwidth]{\detokenize{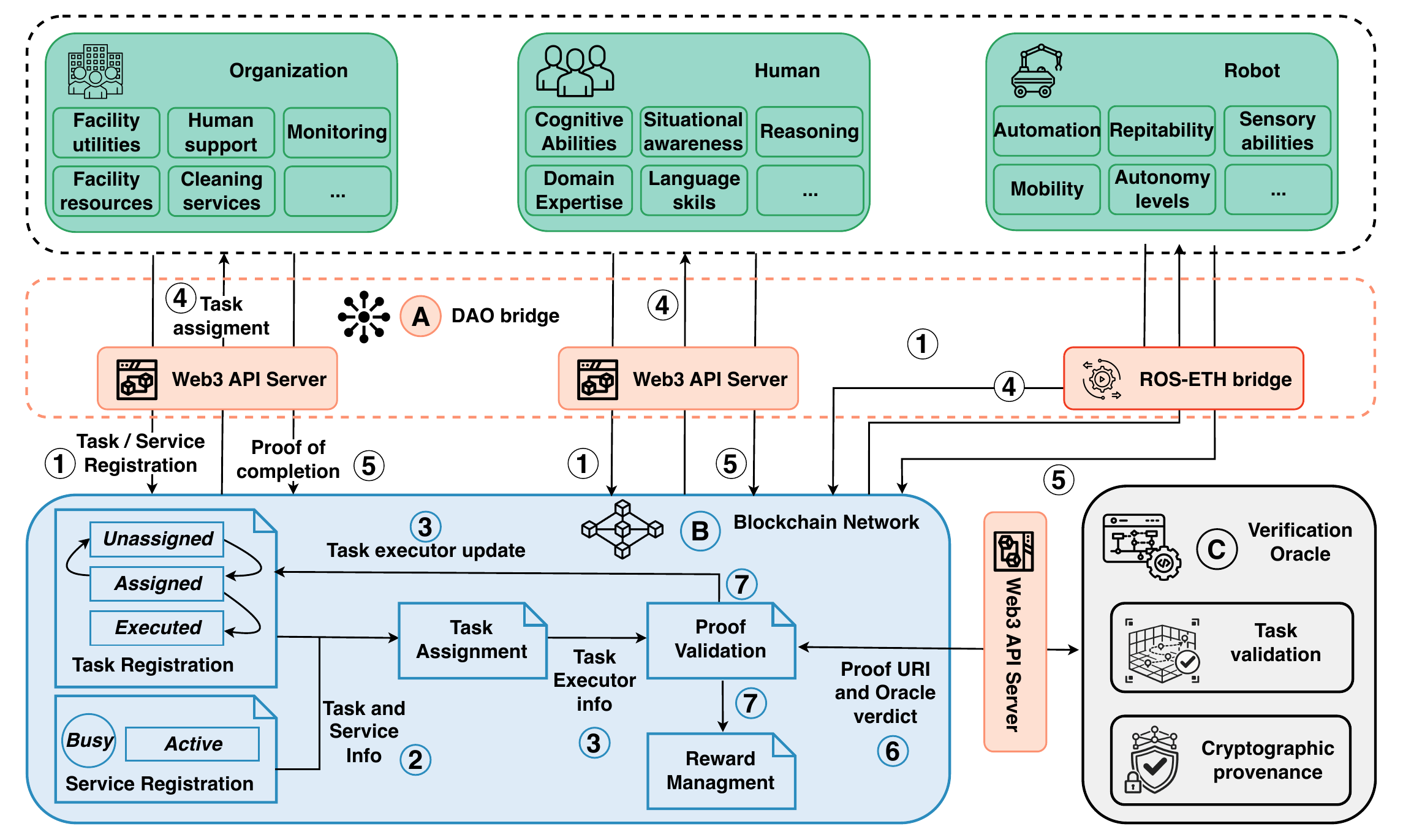}}
  \caption{\textbf{Illustration of the interaction between humans, robots, and organizations by using blockchain-based SCs and proof-verification mechanisms}. The proposed framework is composed of: A) DAO bridge: an interface that gives organizations, humans, and robots a common tool to publish tasks and services, receive assignments, and manage rewards. B) Blockchain network: a platform that defines the rules of the operation of the organization, and C) Verification oracle: proof-verification entity for actions performed in the physical world}
\vspace{-15pt}
  \label{fig:system}
\end{figure*}

Recent research has proposed robots as economic agents that transact with humans and other robots via blockchain technology. For instance, the authors of Robonomics~\cite{danilov2018robonomics} introduced liability contracts to validate robot tasks on Ethereum, while~\cite{kapitonov2021raas} outlined decentralized marketplaces for robotic capabilities. Industry initiatives such as OpenMind’s FABRIC~\cite{openmind2023fabric}, Symbiotic’s shared security framework~\cite{symbiotic2024sharedsecurity}, and the peaq framework for the machine economy~\cite{peaq_machine_economy} extend this concept by envisioning coordination layers and staking based safety guarantees. These projects highlight the industrial potential of decentralized robotic organizations, yet they lack actionable, task-oriented templates that enable service robots to be integrated into organizational workflows.

In response to this, we ground our approach in service robot tasks, which are already common in many organizations (e.g., cleaning~\cite{vanderloos2020service} and logistical~\cite{prassler2010cleaning}) and are expected to become even more integrated in human societies as robotic assistants~\cite{nvidiaPhysicalAI, gemini_robotics_2025}. Ensuring extended robot autonomy that maintains trust, accountability, and explainability is critical in these domains~\cite{srinivas2021survey}, since autonomous decisions directly affect human safety and operational reliability. Without mechanisms that make actions verifiable and decisions interpretable, organizations can not guarantee compliance, traceability, or confidence in the robot’s behavior. Prior work has explored trust and transparency primarily through behavioral design~\cite{fernandes2019robotchain} and interface development~\cite{zhang2022rosethereum}, while industrial deployments have emphasized safety and compliance~\cite{kallweit2015ros}. In contrast, our work focuses on how economic incentives and decentralized organizational structures can extend autonomy while maintaining verifiable trust. By embedding service robots into DAOs, we aim to create systems where robots can not only execute tasks reliably but also engage in verifiable agreements, payments, and organizational decision making---capabilities that traditional autonomy frameworks do not provide. 

\vspace{-15pt}
\section{RODEO design}
\subsection{Problem Definition}
The integration of service robots into an organization presents challenges that are hard to solve with current architectures~\cite{robot-fleet}. To begin with, organizations today deploy robots from different vendors, each specialized in automating specific tasks (e.g., cleaning, delivery, security). These deployments rely on proprietary centralized management systems that do not communicate with each other~\cite{decentralized-robots}. Thus, organizations are forced to manage various robotic systems with complex hardware and software heterogeneity. While ROS~\cite{288} provides hardware and software abstraction, it does not offer a framework for robots to express their functional capabilities to the organization. Consequently, robots can not receive dynamic task assignments based on real-time needs nor can they independently proof that a task was successfully completed.

This inability to independently verify actions makes establishing trust and liability even more challenging, as the operational source of truth remains obscure~\cite{Winfield2017}. When a robot fails or causes damage, the operational logs are typically stored on proprietary servers or the robot’s local disk. These logs are often mutable, meaning they can be altered or deleted.

\subsection{System Architecture}
Fig.~\ref{fig:system} illustrates RODEO, a blockchain-based framework where an organization, robots, and humans (green box) create tasks, advertise services, and receive matched assignments. First, the organization  provides the operational environment, managing resources such as power, maintenance, storage, and security. Second, humans serve as operators and auditors who contribute situational awareness, technical expertise, and reasoning capabilities. Their participation covers activities ranging from hardware repair and software maintenance to supervision, calibration, and decision support. Finally, robots act as autonomous agents that execute physical tasks including cleaning, transport, inspection, and security patrols. In this research, robots can advertise their services, accept assignments, and submit verifiable proofs of task completion, enabling accountable collaboration between humans and robots within an organization. The RODEO framework is organized into three primary architectural building blocks: \emph{A)} DAO bridge that facilitates participant (e.g., organization, humans, robots) interaction, \emph{B)} Blockchain network that implements the matching, assignment, proof verification, and settlement logic and \emph{C)} Verification oracle that validates evidence for successful task execution.

\noindent \textit{A) DAO bridge:}
The DAO bridge (red box) serves as the interface layer between participants in the blockchain network. It provides organizations, humans, and robots with a common tool to publish tasks and services, receive assignments, manage rewards, and observe state updates. In contrast with typical DAO tooling~\cite{Lustenbergeretal2025}, the DAO bridge proposed in this work offers a programmable API that robots can call directly, enabling them to function as autonomous participants within the organization. In other words, prior work (see Sec.~\ref{sec:soa}) links robots to blockchains mainly for logging or low level access, rather than for full organizational participation. The DAO bridge in this paper is designed as a programmable interface that allows robots to directly broadcast their operational needs and advertise their specific capabilities to the rest of the participants.

\noindent \textit{B) Blockchain network:}
The blockchain network (blue box) forms the foundation of RODEO's decentralized operations specifically addressing the critical issues of trust, liability, and incentivized behavior. To resolve the problem of trust and liability, the blockchain network functions as a transparent, tamper-proof ledger that records participants interactions, task completion proofs and robot data in a way that can not be altered by manufacturers or individual participants. The concept of DAO replaces centralized control with an active framework of rules, guaranteeing vendor-agnostic interoperability, trustless liability, and native economic autonomy for the robots. This subsystem utilizes the programmability of the blockchain network to deploy Smart Contracts (SC) in order to define the specific rules on how the organization is governed. In this research, one of these rules is the use of ERC20 escrows to reward participants for the correct execution of tasks in the organization. 

In the workflow shown in Fig.~\ref{fig:system}, participants (e.g., organization, robots, humans) interact with SCs to define task and service registration~\circled[0.1pt]{1}. During this stage, the task creator stakes an escrow representing the payment for task execution. The assignment contract then retrieves these definitions~\circled[0.1pt]{2} and applies a matching policy to pair pending tasks with suitable services. Once matched, the SC updates the task status, records the executor’s wallet address~\circled[0.1pt]{3}, and notifies the service owner~\circled[0.1pt]{4}. After task completion, the owner submits cryptographic proof to the validation SC~\circled[0.1pt]{5}, which triggers a verification oracle to evaluate the evidence and post an on-chain verdict~\circled[0.1pt]{6}. Finally, the reward management contract (i.e., ERC20 escrow) automates settlement~\circled[0.1pt]{7}, unlocking funds for the owner upon success, or refunding the creator and resetting the task upon failure. This ensures rewards are guaranteed rather than voluntary. Rejected or expired tasks are re-queued with original timestamps, while repeated failures may incur penalties.

\noindent \textit{C) Verification oracle:}
The verification oracle (gray box) directly addresses the trust and liability problem by providing an auditable source of truth. This subsystem ensures that reward settlement is based on verified physical performance. The oracle performs verification through two stages: \emph{i)} Cryptographic provenance: it validates that the task execution proof is legitimate and has not been manipulated. This introduces trust in the system ensuring that only trustworthy participants can operate in the decentralized organization.  \emph{ii)} Task validation: Emulation-based verification by replaying the robot or human data proof in a simulator (e.g., Gazebo). This allows the system to verify that the robots physical actions (e.g., paths, sensor data), actually met the task specifications before the on-chain verdict triggers the release of the escrows. 

\begin{figure*}[!t]
    \centering
    \includegraphics[width=\textwidth]{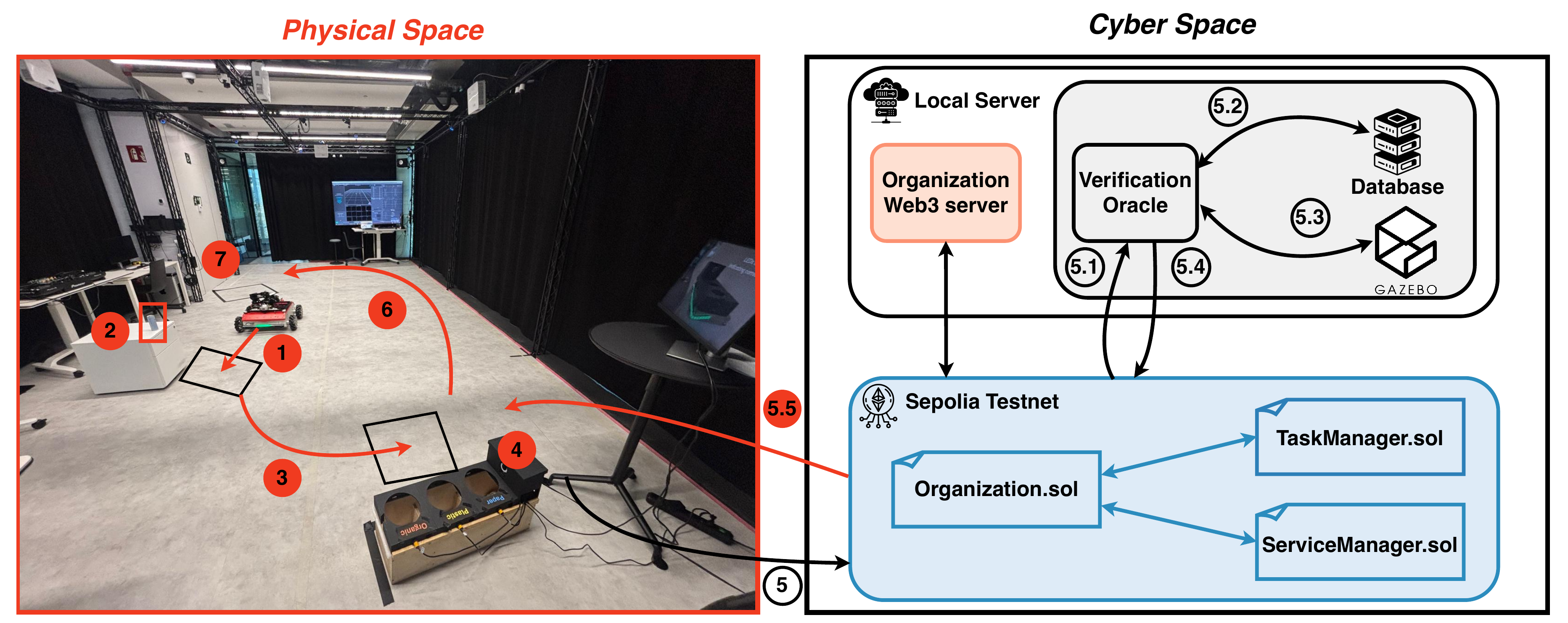}
    \caption{Cleaning task scenario. The figure depicts the physical and cyber sections of this use case. The left section shows the physical space, where a mobile robot equipped with a robotic arm performs the waste collection sequence (steps 1–5) followed by the battery charging process (steps 6 and 7). The right section represents the cyber space, showing the deployment of smart contracts on the public Sepolia testnet and the operation of the verification oracle implemented on a local server connected to the Web3 interface.}
    \vspace{-15pt}
    \label{fig:experimental-scenario}
\end{figure*}

\section{Use case: cleaning task}

\subsection{Experimental Scenario}
To evaluate RODEO we designed a cleaning task scenario\footnote{Demonstration video of RODEO: \url{https://youtu.be/L5voOWKFLzk}} in a university lab (Fig.~\ref{fig:experimental-scenario}). The main participants were an organization (managing tasks and services) and a mobile robot with a manipulator. Two services were established: waste disposal (provided by the robot) and battery charging (provided by the organization). In this scenario, when trash is found in the lab (normally left by the students), the organization creates a waste disposal task, and the system assigns it to an available robot. Then, the robot navigates to the trash (1), grasps the item (2), drives to the bin (3), deposits it in the bin (4), and submits a rosbag file as proof (5). Once the proof is validated the robot receives 100 tokens as a reward. This proof-submission mechanism enforces robot liability. If the robot fails to complete the task or incorrectly disposes of the item, the rosbag proof and the SC escrow ensure the organization is not only refunded, but possesses a verifiable audit trail to determine fault and potentially penalize the robot. When the battery of the robot is low (e.g., after completing several waste disposal tasks), the robot autonomously creates a charging task, moves to the charger (6), and completes a battery recharge cycle. In that moment, the organization submits an energy consumption proof for settlement (7). A completed charging session costs the robot 200 tokens. In the absence of tasks, the robot performs random exploration.

\subsubsection{Physical Space}
The left side of Fig.~\ref{fig:experimental-scenario} shows the lab where experiments took place. This lab is provisioned with an OptiTrack motion capture system, which provides precise position and orientation data for objects within a capture area. This system functions as a practical source of ground truth during experiments, publishing poses into a single ROS TF tree. The same global frame registers the charging dock, tagged waste items, and a smart bin named iTrash~\cite{iTrash}. iTrash is a waste management system that utilizes computer vision for automated sorting. In our setup, iTrash functions as a smart trash bin, providing execution confirmation for cleaning tasks. Step (4) on Fig~\ref{fig:experimental-scenario} shows iTrash in our lab where the robot presents a grasped item to the camera, queries a FastAPI service for classification, and receives a completion flag after dropping the item in the correct bin. This classification and completion response are recorded in a rosbag file, which serves as the on-chain proof for task execution (5), validated by the proof verification oracle.

We evaluated RODEO using a Husarion Panther~\footnote{https://husarion.com/manuals/panther/} mobile base with a Trossen ViperX 300S~\footnote{https://www.trossenrobotics.com/viperx-300} manipulator and a parallel gripper, enabling it to perform tasks such as grasping and depositing items. The robot runs ROS Noetic and uses the standard navigation and manipulation stack for its operations within the lab space. 

\subsubsection{Cyber Space}
\label{sssec:cyber_space}
The cyber space section, illustrated at the right side in Fig.~\ref{fig:experimental-scenario}, orchestrates the decentralized operations and economic incentives of the RODEO framework, primarily through a public blockchain (Sepolia Testnet) and a local server.

The prototype of RODEO was deployed on the Ethereum Sepolia Testnet due to its suitability for testing. To facilitate incentives and accountable coordination, we deployed an ERC20 compatible token named IEC
\footnote{The deployed IEC Token Smart Contract can be found  \href{https://sepolia.etherscan.io/address/0x5D64497a14a5FD35A96F4B8c5AA6603DB56E913C}{\texttt{here}}} 
that was used for rewards in our experimentation. For this proof-of-concept, IEC functions strictly as a internal utility token with no external fiat value. Upon SC deployment, a fixed supply was pre-minted, with 2,000 IEC pre-allocated to both the organization and the robot's wallet to seed initial operations. 

Three SCs coded using the Solidity programming language run in the blockchain network:

\begin{itemize}
    \item \texttt{Organization.sol
    \footnote{The deployed Organization Smart Contract can be found \href{https://sepolia.etherscan.io/address/0xF204681dc989386646b542649E6b1B12De468e2F}{\texttt{here}}}
    }: Orchestrates task to service matching and settlement. It implements that Task Assignment functionality of the blockchain network (see Fig~\ref{fig:system}).
    \item \texttt{TaskManager.sol
    \footnote{The deployed Task Manager Smart Contract can be found \href{https://sepolia.etherscan.io/address/0x6F0B8122bCfaA2CB8E838458662cDdec74673399}{\texttt{here}}}
    }: Stores task records, tracks tasks states and manages IEC escrow for task creators. It implements the Task Registration, Proof Validation and Reward Management functionality of the blockchain network.
    \item \texttt{ServiceManager.sol
    \footnote{The deployed Service Manager Smart Contract can be found \href{https://sepolia.etherscan.io/address/0x7fa7c9651A7da4d342394026796f35Df4E10bF5F}{\texttt{here}}}
    }:  Stores service records and tracks service states. It implements the Service Registration functionality of the blockchain network.
\end{itemize}

\begin{figure}[t]
    \centering
    \includegraphics[width=\linewidth]{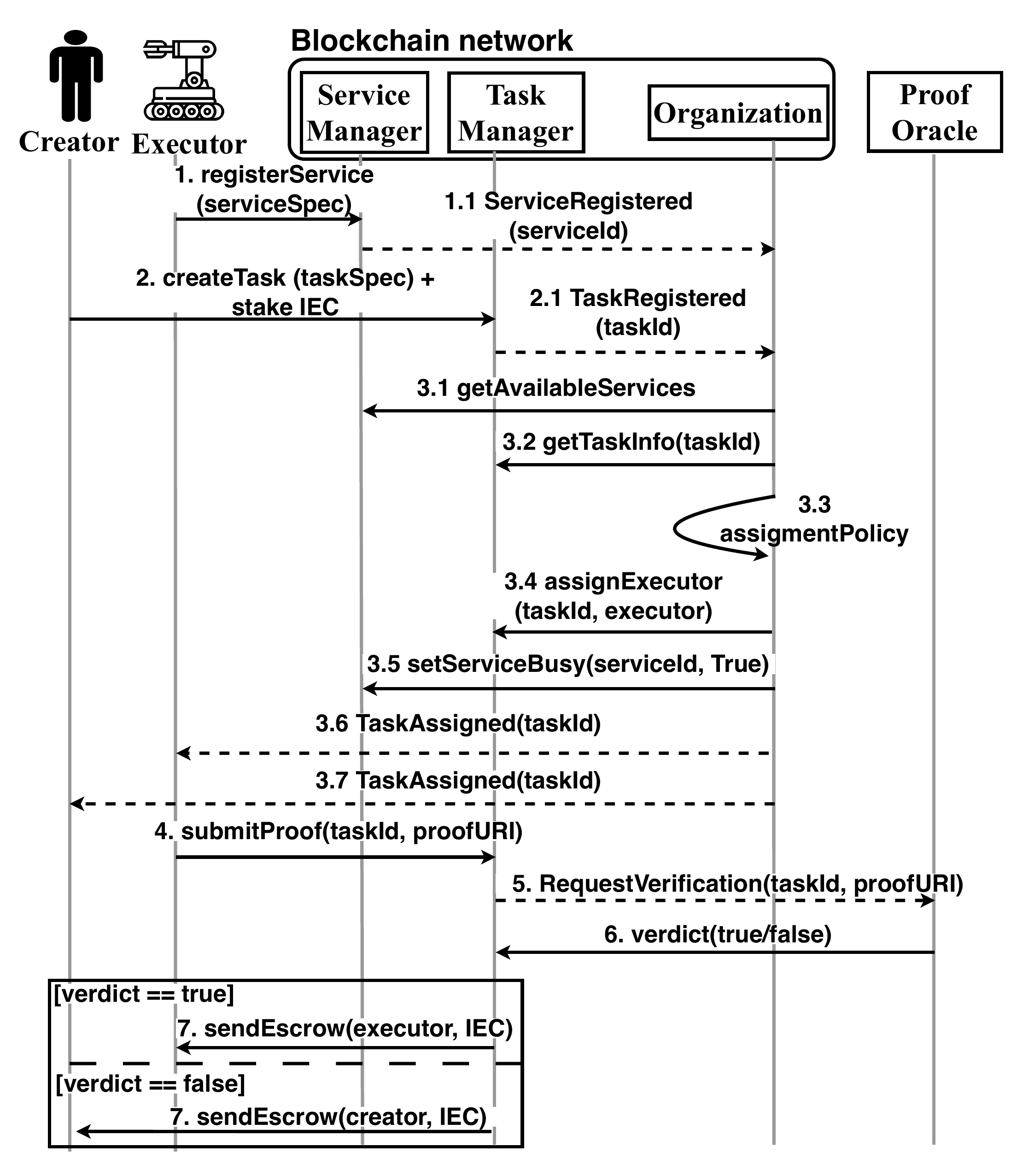}
    \caption{Sequence diagram of the on-chain workflow across \texttt{ServiceManager.sol}, \texttt{TaskManager.sol}, \texttt{Organization.sol}, and the proof-\textbf{verification} oracle.}
    \label{fig:contract-workflow}
\end{figure}

The interactions between the three SCs are summarized in Fig~\ref{fig:contract-workflow}. Participants first publish capabilities through \texttt{ServiceManager.sol} (step 1) and register task execution requests through \texttt{TaskManager.sol} (step 2). Task creation escrows the promised reward in IEC, thereby committing funds prior to execution. Both registrations emit events that are consumed by \texttt{Organization.sol} (step 1.1 and 2.1), which triggers the organization’s assignment policy. Once an event is received, the \texttt{Organization.sol} requests the detailed task information and list of all the available services in the system (step 3.1 and 3.2). Based on this information the \texttt{Organization.sol} contract attempts to map the task requirements to eligible services (step 3.3). If the assignment policy matches a given task to a service, the selected task is updated in the \texttt{TaskManager.sol} with the task executor address (step 3.4) and the corresponding service state is updated in \texttt{ServiceManager.sol} from Active to Busy (step 3.5). The resulting assignment is broadcast via events to notify the task creator and executor (step 3.6 and 3.7). After performing the task, the executor submits a proof reference (task identifier and proof location) to \texttt{TaskManager.sol} (step 4), which triggers the proof-verification oracle and awaits a verdict (step 5). A positive verdict releases the escrowed IEC to the executor (step 6 and step 7), whereas a negative verdict refunds the task creator, after which \texttt{Organization.sol} and/or \texttt{ServiceManager.sol} can restore service availability for subsequent assignments.

For what concerns the proof oracle, the local server hosts a prototype implemented in Python. Upon proof submission (5.1), the oracle inspects its type:

\begin{itemize}
    \item \textbf{rosbag proofs (Cleaning)}: The oracle downloads (5.2) and replays the rosbag in Gazebo, verifying the robot's base trajectory (pick and place stops), manipulator pose changes, gripper release at the bin, and successful iTrash message (5.3).
    \item \textbf{Log Proofs (Charging)}: The oracle downloads and analyses the energy consumption logs to compute total expenditure and checks compliance with the 200 IEC reward policy.
\end{itemize}

To favor reproducibility, we make the oracle implementation, RODEO SCs and ROS-ETH bridge available\footnote{\url{https://github.com/IERoboticsAILab/RODEO}} to other researchers and developers to replicate our setup, fostering transparency and encouraging future real-world experimental studies of robotic DAOs.

\subsection{Experiments and data collection}
We ran our cleaning task scenario for three consecutive days in the space depicted in Fig.~\ref{fig:experimental-scenario}. Each time trash was left in the lab, the organization created a new waste disposal task using a web GUI. Then, this task was assigned to an available robot. For successful execution of the task the robot was rewarded with 100 IEC. When no task was active, the robot executed a random walk coverage while waiting for assignment. When the robot's battery reached a low threshold (i.e., reached 50\% battery level), the robot created a charging task, navigated to the charging area and completed the charging session. For each charging session the robot paid a fixed electricity tariff of 200 IEC to the organization. During the experimentation phase we completed a total of 59 tasks where 8 were battery charging tasks created by the robot and 51 were waste disposal tasks created by the organization.

All interactions were recorded on chain through SC events and transaction receipts. From these logs we computed metrics, including average task execution time from assignment to proof submission and average proof validation time from proof submission to oracle verdict We also tracked the robot wallet balance over time, the number and duration of charging sessions, and the interval between charges. 

\section{Results}
\begin{figure}[tbh]
  \centering
  \includegraphics[width=\linewidth]{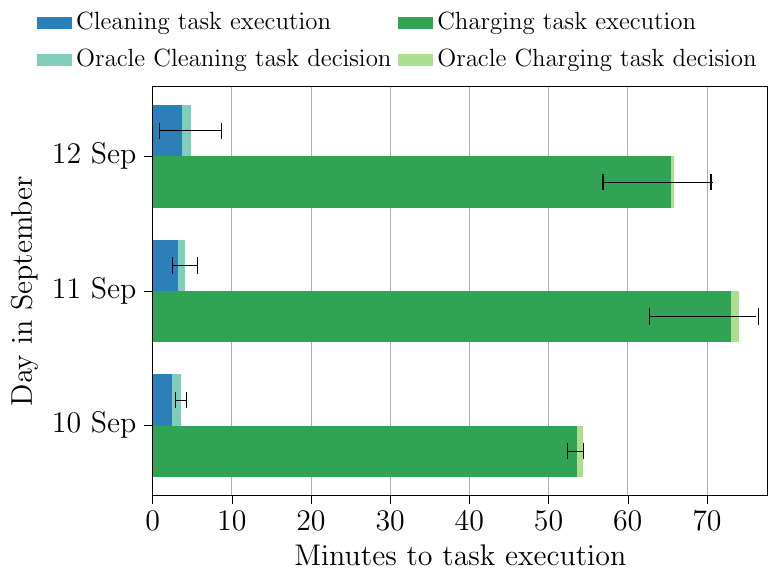}
  \caption{Completion times (in minutes) for task and proof creation. Bar plots show the median completion times for the cleaning (blue) and charging (green) tasks. Light colors show the median time of proof creation after the corresponding task was completed.}
  \vspace{-10pt}
  \label{fig:task_completion_times}
\end{figure}

\begin{figure*}[t!]
  \centering
  \includegraphics[width=\textwidth]{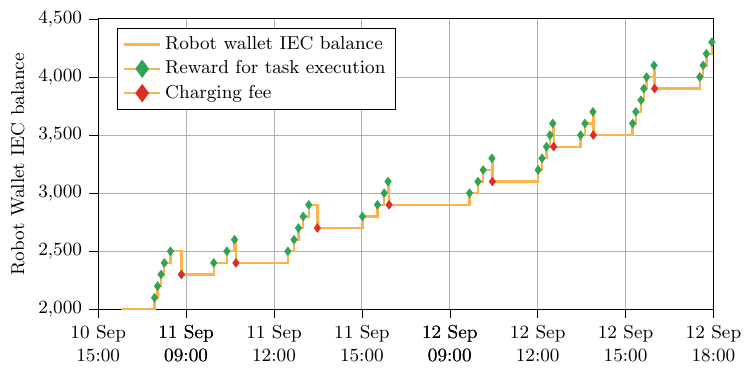}
  \caption{Balance of the robot's wallet (in IEC) from September $10^{th}$-$12^{th}$ 2025. In the figure, green diamonds represent transactions from the organization to the robot as reward for successful execution of the cleaning task. Red diamonds represent the purchase of electricity for battery charging. Finally, the blue line represents the electricity price in \euro{}/kWh for the duration of the experiment.}
  \vspace{-15pt}
  \label{fig:wallet_price}
\end{figure*}
Fig.~\ref{fig:task_completion_times} shows the median completion time for the two task categories (cleaning and charging), broken down into two critical stages: task execution (i.e., the time needed for the task to be executed in the physical world) and oracle decision (i.e., the time needed for the proof to be validated). From Fig.~\ref{fig:task_completion_times}, the charging task (green bar) dominates the overall runtime. Across the three evaluation days (10-12 Sep), the robot’s battery charging time was 53.6, 73.0, and 65.5 minutes, with a median of 65.5 minutes. The corresponding oracle verification time (light green segment) was 0.8, 1.0, and 0.3 minutes, yielding median charging-task completion time of 65.8 minutes. For the cleaning task (blue bar), the task median execution time was 3.2 minutes (range: 2.4–3.7). The oracle median decision time was 1.1 minutes (range: 0.9–1.2), resulting in total cleaning-task completion time of 4.1 minutes (range: 3.6–4.8). This interval includes navigation to the trash, grasping the item, depositing it in the bin, and uploading the proof. The oracle time reflects replaying the rosbag at 3× speed while verifying the execution checkpoints.

Fig.~\ref{fig:wallet_price} shows the evolution of the robot wallet
across the duration of the experiment. The yellow line shows the robot wallet balance in IEC. The drops (red diamonds) in the yellow line mark charging events and flat segments indicate inactivity during night hours. Fig.~\ref{fig:wallet_price} shows that the organization seeded the robot wallet with 2000 IEC, and the balance increased to 4100 IEC through task execution in the lab across three days, which means the robot could return the initial stake and still retain an additional 2100 IEC to sustain further operation. The timing of the balance drops aligns with battery charging during working hours, while the flat balance during the night coincide with periods when the platform was not active due to the lack of students and staff at the university. 

\begin{figure}[tbh]
  \centering
  \includegraphics[width=\linewidth]{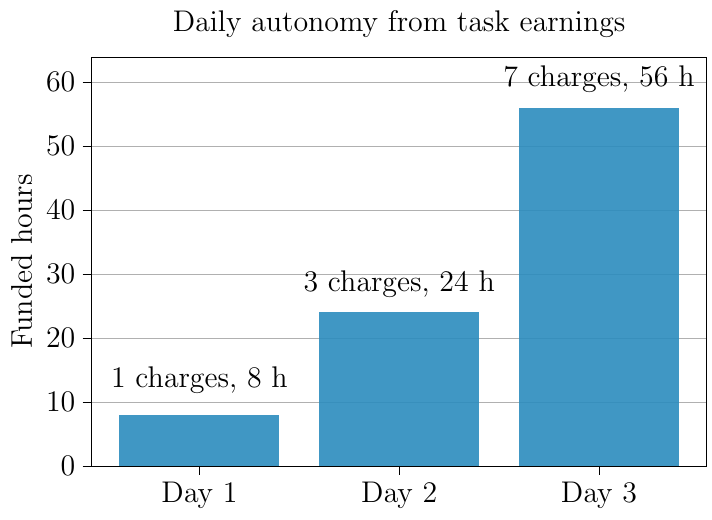}
  \caption{Daily extended operating hours funded by the robot's task earnings. Each bar represents the autonomy gained from that specific end-of-day balance, summing to a total of 11 charges (88 hours) over the three-day period.}
  \vspace{-15pt}
  \label{fig:autonomy_daily}
\end{figure}

Finally, Fig~\ref{fig:autonomy_daily} illustrates how daily task earnings, reinvested as battery charge can extend the robot operating hours. A single charge allows 8 hours of extended operation. The funded hours were calculated based on the robots earnings in the end of each day. For example, on day one, the robot completed 5 tasks (+ 500 IEC) and charged its battery once (- 200 IEC), resulting robot wallet balance of 300 IEC. Reinvesting this 300 IEC in 1 additional charge will allow the robot to extend its operation time for 8 hours. The funded hours increased on day two to 24 hours and on day three to 56, corresponding to three, and seven charging cycles covered, respectively. The robot can sustain activity by investing earnings into electricity, summing to 11 additional charges and it can remain operational for 88 hours while awaiting for new task creation.

\vspace{-0.5pt}
\section{Discussion}

The aim of RODEO is to provide a baseline to real-world experiments in robotic DAOs. By using RODEO, researchers can now explore autonomous organizational designs, new business models, and real-world experimentation with economic and verification mechanisms. Our results demonstrate that a mobile robot can successfully execute a cleaning task, use the ROS-ETH bridge to interact with the blockchain (e.g., submit verifiable proofs of its work), and get compensated in tokens, and used those earnings to purchase required services (i.e., battery charging). Thus extending its autonomy.

Nevertheless, several problems remain unsolved before incentive-based autonomous organizations can be realized. While our current implementation utilizes a first-come-first-served matching and single-token escrow for simplicity, these mechanisms do not sufficiently address advanced DAO governance concerns such as collusion~\cite{tamai2024dao} or Sybil resistance~\cite{siddarth2020who}. Successful task execution hinges on a robust pricing and governance model to prevent economic imbalance. Literature indicates that simple token-weighted systems tend to centralize quickly~\cite{fritsch2024analyzing},~\cite{tamai2024dao}. Without robust mechanisms, we risk creating an organization dominated by a small group of highly efficient participants. To counter this, future iterations of RODEO must move beyond simple market pricing toward reputation-based membership, quadratic voting for task allocation, or stake-slashing mechanisms to resolve disputes.

\subsection{Limitations and future work}
In our experiments, we considered a single robot that is sufficient for validating the ROS-ETH bridge and blockchain network component, but does not allow to investigate the impact of RODEO's economic incentives in a human-robot interactions. For example, in a multi-robot deployment the robot could post tasks(e.g., upgrade my hardware, clean my wheels), incentivizing humans to handle the robot carefully since its continued operation sustains future revenue opportunities. As a future work, we will extend the experimental scenario to include multiple robots and, human participants. This experimental setup will allow us to test if token-based incentives can measurably improve human-robot collaborations.

Another limitation of our study is that RODEO was evaluated using only two distinct tasks and corresponding services. This minimal setup provided insufficient economic complexity to rigorously evaluate the interplay between different pricing models and re-investment strategies. To address this, our future research will leverage large-scale simulations using Gazebo and PyRoboSim to model the organization's evolution under a wide variety of conditions, including fluctuating numbers of robots, diverse task availability, and pricing models. 

As we scale the number of robots in the organization, the integration of a blockchain network can introduce operational trade-offs regarding gas prices, network latency, and system complexity. As the number of robots and tasks grow the organization we will be exposed to fluctuating transaction fees and block confirmation delays, which could effect the real-time robot responsiveness. This operational overload underscores the necessity for a careful selection of compatible Ethereum-based platform (e.g., leveraging Layer-2 scaling or sidechain integrations) to ensure that RODEO remains viable for multi-robot organizations.

Finally, our PoC implemented a simple oracle and rosbag-based proof submission to prioritize the evaluation of RODEO as a coordination framework. However, this approach remains vulnerable to on robot spoofed messages or replayed telemetry. To mitigate this, future research will leverage hardware-backed signing or tamper-evident "sealed rosbag", so we can ensure that the data submitted to the DAO has not been modified.

\section{Conclusions}
Designing organizations composed of service robots that are capable of acting, generating revenue, and reinvesting it into their own sustainability remains a significant research challenge. While proof-of-concepts of economically autonomous robots have recently been demonstrated, scaling such concepts to organizations is non-trivial. In this work, we presented RODEO, a blockchain-based framework that allows service robots to operate within decentralized organizations, interact with other agents, and autonomously manage their economic lifecycle. Through an in-lab experimentation, we demonstrated a mobile robot that performs cleaning tasks, submits verifiable execution proofs, receives token-based compensation, and pays for electricity to continue operating. All interactions were recorded and settled on a public blockchain using SC and an off-chain oracle, ensuring that each task execution and payment was auditable. Over the course of the experiment, the robot maintained a positive balance in its wallet, reinvested its earnings into operational resources, and demonstrated the ability to remain functional without external intervention. These results demonstrate that robots can be embedded in programmable institutions where their work verifiable and payable.




\section*{ACKNOWLEDGMENT}
This work was supported by project PID2023-152334OA-I00, funded by MCIU/AEI/10.13039/501100011033 and by FEDER, EU. It was also supported by the Ramón y Cajal fellowship RYC2023-043120-I, funded by MCIU/AEI/10.13039/501100011033 and by FSE+. Additional funding was provided by the 2024 Leonardo Grant for Scientific Research and Cultural Creation (LEO24-1-12086-CCD-CIA-1) from the BBVA Foundation. The BBVA Foundation accepts no responsibility for the opinions, statements and contents included in this publication, which are entirely the responsibility of the authors.


\bibliography{references}
\bibliographystyle{IEEEtran}
\end{document}